\documentclass[10pt,conference]{IEEEtran}

\usepackage{cite}

\ifCLASSINFOpdf
   \usepackage[pdftex]{graphicx}
   \graphicspath{{figs/}}
   \DeclareGraphicsExtensions{.pdf,.jpeg,.png}
\else
   \usepackage[dvips]{graphicx}
   \graphicspath{{../figs/}}
   \DeclareGraphicsExtensions{.eps}
\fi
\usepackage[cmex10]{amsmath}
\usepackage{amsthm}

\usepackage{algorithmic}

\usepackage{array}

\ifCLASSOPTIONcompsoc
  \usepackage[caption=false,font=normalsize,labelfont=sf,textfont=sf]{subfig}
\else
  \usepackage[caption=false,font=footnotesize]{subfig}
\fi
\usepackage{url}

\newif\iffinal
\finaltrue
\newcommand{\cmtid}{99999}

\iffinal
\else
\usepackage[switch]{lineno}
\fi

\usepackage{hyperref}
\usepackage{booktabs}
\usepackage{multirow}
\usepackage{float}
\usepackage{amssymb}
\usepackage{latexsym}

\begin{document}
%
\title{Toward unlabeled multi-view 3D pedestrian detection by generalizable AI: techniques and performance analysis}


\iffinal



 
\author{\IEEEauthorblockN{João Paulo Lima\IEEEauthorrefmark{1}\IEEEauthorrefmark{2},
Diego Thomas\IEEEauthorrefmark{3},
Hideaki Uchiyama\IEEEauthorrefmark{4} and 
Veronica Teichrieb\IEEEauthorrefmark{2}}
\IEEEauthorblockA{\IEEEauthorrefmark{1}Departamento de Computação, Universidade Federal Rural de Pernambuco, Recife, PE, Brazil, Email: joao.mlima@ufrpe.br}
\IEEEauthorblockA{\IEEEauthorrefmark{2}Voxar Labs - Centro de Informática, Universidade Federal de Pernambuco, Recife, PE, Brazil}
\IEEEauthorblockA{\IEEEauthorrefmark{3}Faculty of Information Science and Electrical Engineering, Kyushu University, Fukuoka, Japan}
\IEEEauthorblockA{\IEEEauthorrefmark{4}NARA Institute of Science and Technology (NAIST), Nara, Japan}}

\else
  \author{Sibgrapi paper ID: \cmtid \\ }
  \linenumbers
\fi

\maketitle

\begin{abstract}
We unveil how generalizable AI can be used to improve multi-view 3D pedestrian detection in unlabeled target scenes.
One way to increase generalization to new scenes is to automatically label target data, which can then be used for training a detector model.
In this context, we investigate two approaches for automatically labeling target data: pseudo-labeling using a supervised detector and automatic labeling using an untrained detector (that can be applied out of the box without any training).
We adopt a training framework for optimizing detector models using automatic labeling procedures.
This framework encompasses different training sets/modes and multi-round automatic labeling strategies.
We conduct our analyses on the publicly-available WILDTRACK and MultiviewX datasets.
We show that, by using the automatic labeling approach based on an untrained detector, we can obtain superior results than directly using the untrained detector or a detector trained with an existing labeled source dataset.
It achieved a MODA about $4\%$ and $1\%$ better than the best existing unlabeled method when using WILDTRACK and MultiviewX as target datasets, respectively.
\end{abstract}


\IEEEpeerreviewmaketitle

\section{Introduction}

Detecting pedestrians is a persistent challenge in smart cities, surveillance, monitoring, autonomous driving, and robotics, among other fields.
There is growing interest in estimating the 3D location of pedestrians as it facilitates georeferencing people in the 3D environment.
Robust and accurate 3D pedestrian detection is essential to develop sustainable smart cities that are aware of the uses of public space.
Pedestrians' 3D location can be obtained from a single camera using 3D monocular detectors, but they do not handle occlusions very well.
By using multiple cameras, multi-view constraints can be leveraged to improve the accuracy of the estimated 3D pedestrian locations.
Nowadays, it is common for areas to be monitored using multiple monocular cameras with overlapping fields of view, such as security cameras.
This setup facilitates 3D pedestrian detection by exploiting multi-view constraints and better handling occlusions.
Nonetheless, multi-camera 3D pedestrian detection in crowded environments remains a challenging task.

In recent years, AI-based methods have made tremendous progress in the multi-view 3D pedestrian detection field.
These approaches are based on pedestrian occupancy density estimation on the ground level by heatmap regression using deep neural networks~\cite{hou2020multiview, song2021stacked, hou2021multiview, gao2022exploiting, rui20223d, engilberge2023two}.
The key idea is to aggregate multi-view people detection information by applying a feature perspective transform to place pedestrians' ground heatmaps (and later locations) in the same coordinate space.
However, good performance is constrained by data that matches the training datasets.
This consistency requirement is a big issue because the same camera and usage conditions (lighting, weather, etc.) cannot be guaranteed for surveillance, monitoring, and other smart city applications.
The current state-of-the-art methods for detecting pedestrians in 3D using multiple cameras need laborious annotation of ground-truth data from the target scene to achieve the best results.
Consequently, it is desirable to have multi-view 3D pedestrian detection solutions that do not require labeled target scene data~\cite{lima2021generalizable, lima20223d, vora2023bringing}.

Generalizable AI aims to create models to better deal with new scenarios, domains, and tasks~\cite{jiang2021regressive, kim2022unified, arpit2022ensemble}.
One alternative that may help achieve this goal is automatically obtaining target data labels.
For example, pseudo-labeling methods use a model trained with labeled source data to create labels for target data~\cite{lee2013pseudo}.

\begin{figure*}[h]
  \centering  
   \includegraphics[width=\linewidth]{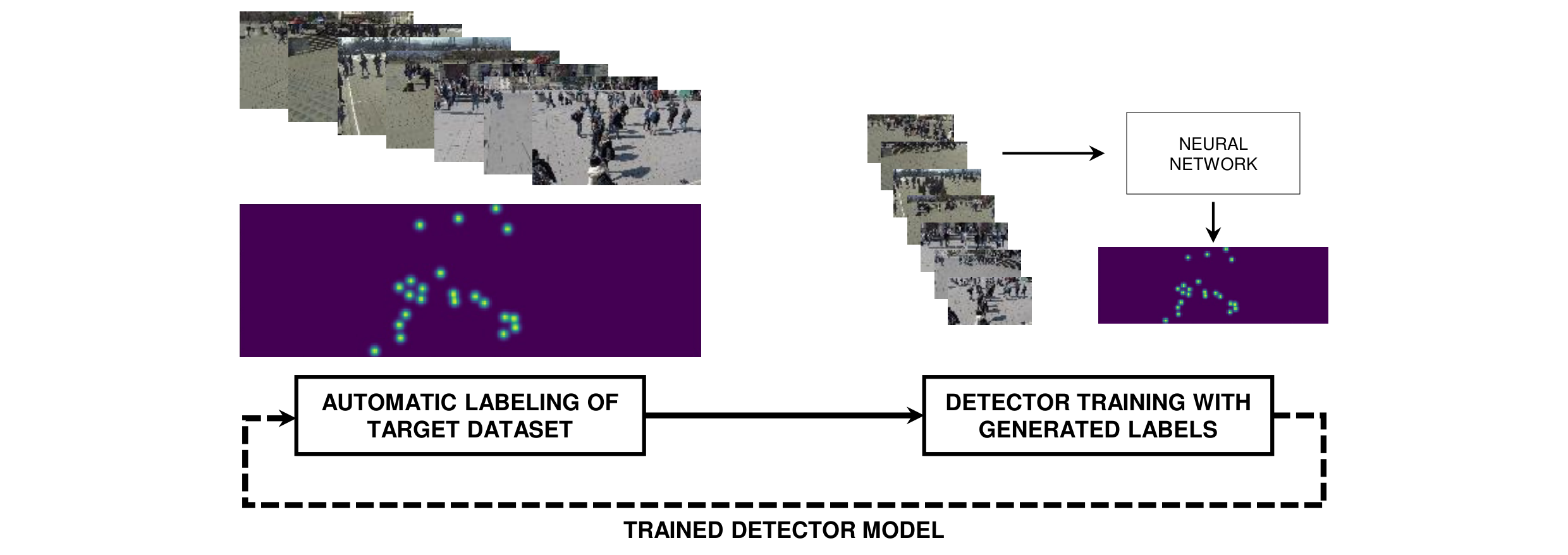}
   \caption{Summary of the approach for unlabeled multi-view 3D pedestrian detection.
   We automatically generate labels for a target dataset.
   Then, we train a detection neural network with generated target labels.
   As optional steps, we can perform multiple rounds of this approach using the trained detector model to regenerate target labels.}
   \label{fig:overview}
\end{figure*}

In this context, our main contribution is to unveil how generalizable AI can improve unlabeled multi-camera 3D pedestrian detection.
To the best of our knowledge, this is the first work to investigate generalizable AI methods in the context of unlabeled multi-view 3D pedestrian detection based on heatmap regression.
Such AI-based approaches aim to make pedestrian localization robust to domain shift, generalizing to various target scene data with varying multi-camera configurations and environmental conditions.
This allows for increasing real-world applicability of pedestrian detection, paving the way for its practical use.
First, we investigate pseudo-labeling of multi-view 3D pedestrian data that employs a supervised detector tailored to occupancy heatmap regression.
Next, we evaluate automatic labeling of multi-view samples that exploits an untrained pedestrian detector.
We adopt a framework for training a multi-view 3D pedestrian detector using automatic labeling procedures.
This framework comprises training set/mode definition and multi-round labeling strategy.
We conduct quantitative and qualitative evaluations of these approaches compared to state-of-the-art unlabeled techniques, showing that some can improve detection performance.

\section{Methods}

Multi-view 3D pedestrian detection methods usually output occupancy heatmaps on the ground plane for each frame.
An occupancy heatmap represents crowd density from a bird’s eye view of the ground plane.
If occupancy heatmap labels are available for a target dataset, we can use them for training a supervised detector model.
We can then apply the trained model to regress occupancy heatmaps from multi-view samples.
Finally, we can extract pedestrian locations from the estimated heatmaps.
We investigate methods that automatically obtain occupancy heatmap labels for target data in Subsection~\ref{sec:auto_label}.
We also describe the details of our training framework in Subsection~\ref{sec:train_framework}.

\subsection{Automatic labeling}
\label{sec:auto_label}

We summarize the automatic labeling approach in Figure~\ref{fig:overview}.
First, we automatically generate occupancy heatmap labels for a given target dataset.
Then, we train a detector using target data and their corresponding generated heatmap labels.
Optionally, we can repeat this process for multiple rounds using pseudo-labeling.
We can define pseudo-labeling as a type of automatic labeling that exploits existing labeled data to create labels for target data.
While automatic labeling methods generally do not require using an existing labeled dataset, pseudo-labeling methods can employ such dataset to obtain better target labels.

We consider two different options for the automatic generation of heatmap labels.
The first one, which we present in Subsection~\ref{sec:pseudo_label}, consists in pseudo-labeling using a supervised detector based on occupancy heatmap regression.
The second one, which we introduce in Subsection~\ref{sec:auto_label_untrained}, consists in automatic labeling using an untrained pedestrian detector, i.e., that can be applied out of the box without any training.



\subsubsection{Pseudo-labeling with supervised detector}
\label{sec:pseudo_label}

The first automatic labeling approach starts by training a pedestrian detector with the supervision of available labeled data for heatmap regression.
Then we apply the trained detector to obtain pseudo-labels for the target dataset.
For example, if we have a source dataset with ground-truth pedestrian locations available, we can use it to train the detector.
Any supervised detector that outputs occupancy heatmaps can be used~\cite{hou2021multiview, engilberge2023two, vora2023bringing}.
Once we finish training a detector model, we use it to infer occupancy heatmaps for target data.
Next, we threshold the obtained heatmaps with a minimum probability value and apply non-maximum suppression (NMS) for extracting pedestrian locations from the filtered heatmaps.
Then we generate occupancy maps from the extracted pedestrian locations.
We define the set of extracted pedestrian locations in a frame $t$ as $\mathcal{D}_t=\{\mathbf{d}^t_i,i\in[1,N]\}$, where $N$ is the number of pedestrians, and $\mathbf{d}^t_i\in\mathbb{R}^2$ is the location of the $i$-th pedestrian in frame $t$.
The value at location $\mathbf{p}\in\mathbb{R}^2$ in the occupancy map $\mathbf{O}_t$ for frame $t$ is defined as

\begin{equation}
\label{eq:occupancy}
\mathbf{O}_t(\mathbf{p})=
\begin{cases}
1, & \mathbf{p}\in\mathcal{D}_t \\ 
0, & \text{otherwise}.
\end{cases}
\end{equation}

Lastly, we filter these maps with a Gaussian kernel to get the final heatmap labels for target data. This way, the occupancy heatmap label $\mathbf{H}^*_t$ for frame $t$ is given by

\begin{equation}
\label{eq:heatmap}
\mathbf{H}^*_t=\mathbf{O}_t\otimes\mathbf{G}(\sigma),
\end{equation}
where $\otimes$ is the convolution operator, and $\mathbf{G}(\sigma)$ is the Gaussian kernel centered at the origin $\mathbf{o}$ with standard deviation $\sigma$, defined as

\begin{equation}
\label{eq:gaussian}
    \mathbf{G}(\mathbf{p},\sigma)=\frac{\exp(-\frac{\left \| \mathbf{p} - \mathbf{o} \right \|^2_2}{2\sigma^2})}{2\pi\sigma^2}.
\end{equation}

\subsubsection{Automatic labeling with untrained detector}
\label{sec:auto_label_untrained}

Unlike the pseudo-labeling method presented in Subsection~\ref{sec:pseudo_label}, the second automatic labeling approach for multi-view 3D pedestrian detection does not require an existing labeled dataset.

If we have a detector available that does not require training, we can directly apply it to obtain labels for a target dataset automatically.
Any untrained detector can be used~\cite{lima2021generalizable, lima20223d}.
First, we estimate pedestrian locations for target data with the untrained detector.
Then, we employ the same procedure in the pseudo-labeling approach for generating heatmap labels from detected pedestrian locations (occupancy map filtered with Gaussian kernel).
We compute occupancy maps from the obtained pedestrian locations with Equation~\ref{eq:occupancy}.
Finally, we get the heatmap labels by filtering these occupancy maps as in Equation~\ref{eq:heatmap} with the Gaussian kernel defined in Equation~\ref{eq:gaussian}.


\subsection{Training framework}
\label{sec:train_framework}

We have two options for the definition of the training set to be used:
\begin{enumerate}  
  \item The training data consist only of automatically labeled target samples;
  \item The training data include labeled source samples and automatically labeled target samples.
\end{enumerate}

Regarding the training mode to be adopted, we can choose between two alternatives:
\begin{enumerate}
  \item Training the detector model from scratch by initializing the model weights with random values;
  \item Fine-tuning a previously trained detector model by initializing the model weights with pretrained values.
\end{enumerate}

Optionally, we can perform additional rounds of pseudo-labeling when using the last two options of training set.
Following this strategy, we use the detector obtained in the last round to regenerate the target dataset labels in each new round.
We can then exploit such labels for retraining the detector in a bootstrap process.

\begin{table*}[h]
\centering
\caption{Performance evaluation of the strategies for automatic labeling.
Upper part: unlabeled results from GMVD~\cite{vora2023bringing} with different combinations of training data, comprising labeled source data (LS), pseudo-labeled target data with GMVD~\cite{vora2023bringing} (PLT), and automatically labeled target data with the detector by Lima et al.~\cite{lima20223d} (ALT) --- best values in bold, higher is better.
Lower part: results from the untrained detector by Lima et al.~\cite{lima20223d} and labeled GMVD~\cite{vora2023bringing} that also uses labeled target data (LT), for comparison.
Right part: results for different source$\rightarrow$target settings.}
\label{tab:train_data}
\begin{tabular}{@{}ll|rrrr|rrrr@{}}
\toprule
\multirow{2}{*}{Detector} & \multirow{2}{*}{Training data} & \multicolumn{4}{c|}{MultiviewX$\rightarrow$WILDTRACK}                                                                  & \multicolumn{4}{c}{WILDTRACK$\rightarrow$MultiviewX}                                                                  \\ \cmidrule(l){3-10} 
                          &                                & \multicolumn{1}{c}{MODA} & \multicolumn{1}{c}{MODP} & \multicolumn{1}{c}{Precision} & \multicolumn{1}{c|}{Recall} & \multicolumn{1}{c}{MODA} & \multicolumn{1}{c}{MODP} & \multicolumn{1}{c}{Precision} & \multicolumn{1}{c}{Recall} \\ \midrule
\multirow{5}{*}{GMVD~\cite{vora2023bringing}}     & LS only                        & 0.690                    & 0.727                    & 0.834                         & \textbf{0.862}              & 0.242                    & 0.682                    & 0.974                         & 0.249                      \\
                          & PLT only             & 0.672                    & 0.729                    & 0.854                         & 0.811                       & 0.212                    & 0.706                    & 0.957                         & 0.222                      \\
                          & ALT only      & \textbf{0.815}           & 0.725                    & 0.955                         & 0.855                       & \textbf{0.723}           & 0.787                    & 0.968                         & \textbf{0.748}             \\
                          & LS + PLT              & 0.745                    & 0.738                    & 0.891                         & 0.849                       & 0.227                    & 0.700                    & 0.988                         & 0.230                      \\
                          & LS + ALT      & 0.753                    & \textbf{0.745}           & \textbf{0.961}                & 0.785                       & 0.709                    & \textbf{0.800}           & \textbf{0.989}                & 0.717                      \\ \midrule
Lima et al.~\cite{lima20223d}               & -                              & 0.778                    & 0.825                    & 0.878                         & 0.903                       & 0.748                    & 0.892                    & 0.977                         & 0.766                      \\
GMVD~\cite{vora2023bringing}                      & LS + LT                        & 0.872                    & 0.756                    & 0.929                         & 0.944                       & 0.840                    & 0.793                    & 0.976                         & 0.861                      \\ \bottomrule
\end{tabular}
\end{table*}

\section{Experiments}
\label{sec:exp}

We assessed the presented approaches in a crowded multi-camera 3D pedestrian detection scenario.
In the subsequent subsections, we provide the specifics of the conducted experiments and the achieved outcomes.

\subsection{Datasets and metrics}

We employed two publicly available datasets captured by multiple cameras with overlapping fields of view.
Both datasets feature intrinsic and extrinsic calibration for every camera and synchronized frames with a resolution of $1920\times1080$.
Ground-truth 3D locations of pedestrians are available for 400 frames at 2 fps.

The first dataset, called the WILDTRACK (WT) dataset\footnote{\url{https://www.epfl.ch/labs/cvlab/data/data-wildtrack/}}~\cite{chavdarova2018wildtrack}, is a challenging dataset captured using seven static cameras (four GoPro Hero 3 and three GoPro Hero 4) in a crowded public open area.
It covers an area of interest measuring $12\times36$ $m^2$, with an average of 23.8 people per frame, 3.74 cameras covering each scene location, and 9,518 annotations.
The camera intrinsic and extrinsic calibration was performed using a publicly available suite\footnote{\url{https://github.com/idiap/multicamera-calibration/}}.

The second dataset we used is the synthetic MultiviewX (MVX) dataset\footnote{\url{https://github.com/hou-yz/MultiviewX/}}~\cite{hou2020multiview}, which was obtained using six virtual static cameras.
It covers an area of interest of $16\times25$ $m^2$, with an average of around 40 people per frame, 4.41 cameras covering each scene location, and 15,494 annotations.
As the dataset is synthetic, intrinsic and extrinsic parameters of each camera were directly set.

We follow the supervised settings of the datasets, using only the last 10\% of the annotated frames for testing.
The evaluation protocol proposed by Chavdarova et al.~\cite{chavdarova2018wildtrack} was adopted, which employs the following metrics: Multiple Object Detection Accuracy (MODA), Multiple Object Detection Precision (MODP), precision, and recall (higher values are better).
The 3D detections are matched to ground truth using Hungarian matching and only if they are within $0.5 m$ of each other.
MODA is deemed the primary performance metric as it accounts for false negatives and false positives.

\subsection{Environment setup}

We used GMVD~\cite{vora2023bringing} as the supervised detector and the method by Lima et al.~\cite{lima20223d} as the untrained detector.
We adopt a train-validation split of 90\%:10\% for source data, a train-validation-test split of 80\%:10\%:10\% for target data, and a batch size of $1$ sample.
We train the neural network for $10$ epochs and select the model with the best validation accuracy throughout training as the final one for testing.
We use the SGD optimizer with a learning rate of $0.0005$, a momentum of $0.9$, a weight decay of $0.0005$, and the one-cycle learning rate scheduler using a maximum learning rate of $0.005$.
We use a ResNet18~\cite{he2016deep} backbone with pre-trained ImageNet~\cite{deng2009imagenet} weights for feature extraction.
The Gaussian kernel for obtaining the heatmap labels has a size of $41\times41$ with $\sigma=5$.
We filter the output occupancy heatmap with a minimum probability of $0.4$, and
then apply NMS on the proposals using a Euclidean distance threshold of $0.5 m$.


\subsection{Quantitative evaluation}

We tested two settings of source$\rightarrow$target datasets: MVX$\rightarrow$WT and WT$\rightarrow$MVX.
First, we evaluated training the GMVD detector ``from scratch'' (with pre-trained ResNet18 weights only) using different training data.
We employed combinations of the following training sets:

\begin{itemize}
  \item \textbf{LS}: labeled source data;
  \item \textbf{LT}: labeled target data; 
  \item \textbf{PLT}: pseudo-labeled target data with GMVD;
  \item \textbf{ALT}: automatically labeled target data with the Lima et al.~\cite{lima20223d} detector.
\end{itemize}

\begin{table*}[h]
\centering
\caption{Performance evaluation of different training modes using the automatic labeling approach with the detector by Lima et al.~\cite{lima20223d} to generate target training data (ALT).
Upper part: unlabeled results from GMVD~\cite{vora2023bringing} with training ``from scratch'' (FS) and with fine-tuning of a model trained with labeled source data (FT) --- best values in bold, higher is better.
Lower part: results from the untrained detector by Lima et al.~\cite{lima20223d} and labeled GMVD~\cite{vora2023bringing} that uses labeled source data (LS) and labeled target data (LT), for comparison.
Right part: results for different source$\rightarrow$target settings.}
\label{tab:train_mode}
\begin{tabular}{@{}lll|rrrr|rrrr@{}}
\toprule
\multirow{2}{*}{Detector} & \multirow{2}{*}{Mode} & \multirow{2}{*}{Training data} & \multicolumn{4}{c|}{MultiviewX$\rightarrow$WILDTRACK}                                                                  & \multicolumn{4}{c}{WILDTRACK$\rightarrow$MultiviewX}                                                                  \\ \cmidrule(l){4-11} 
                          &                                &                                & \multicolumn{1}{c}{MODA} & \multicolumn{1}{c}{MODP} & \multicolumn{1}{c}{Precision} & \multicolumn{1}{c|}{Recall} & \multicolumn{1}{c}{MODA} & \multicolumn{1}{c}{MODP} & \multicolumn{1}{c}{Precision} & \multicolumn{1}{c}{Recall} \\ \midrule
GMVD~\cite{vora2023bringing}                      & FS                 & ALT only      & \textbf{0.815}           & 0.725                    & \textbf{0.955}                & 0.855                       & 0.723                    & 0.787                    & 0.968                         & 0.748                      \\
                          & FT                    & ALT only      & 0.809                    & \textbf{0.736}           & 0.948                         & \textbf{0.856}              & \textbf{0.756}           & \textbf{0.795}           & \textbf{0.984}                & \textbf{0.768}             \\ \midrule
Lima et al.~\cite{lima20223d}               & -                              & -                              & 0.778                    & 0.825                    & 0.878                         & 0.903                       & 0.748                    & 0.892                    & 0.977                         & 0.766                      \\
GMVD~\cite{vora2023bringing}                      & FS                 & LS + LT                        & 0.872                    & 0.756                    & 0.929                         & 0.944                       & 0.840                    & 0.793                    & 0.976                         & 0.861                      \\ \bottomrule
\end{tabular}
\end{table*}

As seen in Table~\ref{tab:train_data}, using PLT only did not bring much improvement to GMVD compared to ``LS only''.
Adding PLT to LS improved almost all GMVD metrics in the MVX$\rightarrow$WT setting.
However, it caused a decrease in MODA and recall in the WT$\rightarrow$MVX setting.
This result can be explained by the fact that the GMVD model trained with WILDTRACK did not generalize well to MultiviewX.
Adding ALT to LS increased almost all GMVD metrics in both source$\rightarrow$target settings.
``GMVD ALT only'' obtained the best results concerning MODA.
As expected, the unlabeled GMVD results are worse than the ones obtained by labeled GMVD (LS + LT), which can be considered a golden standard.
Nevertheless, ``GMVD ALT only'' outperformed the Lima et al.~\cite{lima20223d} detector in the MVX$\rightarrow$WT setting regarding MODA and precision.

\begin{table*}[t!]
\centering
\caption{Performance evaluation of multi-round automatic labeling.
In the first round, we use automatically labeled target data with the approach based on the detector by Lima et al.~\cite{lima20223d} (ALT) for training.
In the remaining rounds, we use pseudo-labeled target data (PLT) with the presented approach for training.
Upper part: unlabeled results from GMVD~\cite{vora2023bringing} with multiple rounds of automatic labeling --- best values in bold, higher is better.
Lower part: results from the untrained detector by Lima et al.~\cite{lima20223d} and labeled GMVD~\cite{vora2023bringing} that uses labeled source data (LS) and labeled target data (LT), for comparison.
Right part: results for different source$\rightarrow$target settings.}
\label{tab:multi_round}
\begin{tabular}{@{}lrl|rrrr|rrrr@{}}
\toprule
\multirow{2}{*}{Detector} & \multicolumn{1}{l}{\multirow{2}{*}{\begin{tabular}[c]{@{}l@{}}\# of \\ AL rounds\end{tabular}}} & \multirow{2}{*}{\begin{tabular}[c]{@{}l@{}}Training\\ data\end{tabular}} & \multicolumn{4}{c|}{MultiviewX$\rightarrow$WILDTRACK}                                                                  & \multicolumn{4}{c}{WILDTRACK$\rightarrow$MultiviewX}                                                                  \\ \cmidrule(l){4-11} 
                          & \multicolumn{1}{l}{}                                                                            &                                                                               & \multicolumn{1}{c}{MODA} & \multicolumn{1}{c}{MODP} & \multicolumn{1}{c}{Precision} & \multicolumn{1}{c|}{Recall} & \multicolumn{1}{c}{MODA} & \multicolumn{1}{c}{MODP} & \multicolumn{1}{c}{Precision} & \multicolumn{1}{c}{Recall} \\ \midrule
GMVD~\cite{vora2023bringing}                      & 0                                                                                               & LS only                                                                           & 0.690                    & 0.727                    & 0.834                         & \textbf{0.862}              & 0.242                    & 0.682                    & 0.974                         & 0.249                      \\
                      & 1                                                                                               & ALT only                                                                           & 0.809                    & 0.736                    & 0.948                         & 0.856                       & \textbf{0.756}                    & 0.795                    & \textbf{0.984}                         & \textbf{0.768}                      \\
                                            & 2                                                                                               & PLT only                                                                           & \textbf{0.817}                    & \textbf{0.746}                    & \textbf{0.962}                         & 0.851                       & 0.737                    & 0.796                    & 0.978                         & 0.754                      \\
                          & 3                                                                                               & PLT only                                                                           & 0.775                         & 0.739                & \textbf{0.962}                              & 0.807                   & 0.733                & \textbf{0.799}                & 0.983                     & 0.746                  \\ \midrule
Lima et al.~\cite{lima20223d}               & -                                                                                               & -                                                                             & 0.778                    & 0.825                    & 0.878                         & 0.903                       & 0.748                    & 0.892                    & 0.977                         & 0.766                      \\
GMVD~\cite{vora2023bringing}                      & -                                                                                               & LS + LT                                                                       & 0.872                    & 0.756                    & 0.929                         & 0.944                       & 0.840                    & 0.793                    & 0.976                         & 0.861                      \\ \bottomrule
\end{tabular}
\end{table*}

In the next experiment, we evaluated ``GMVD ALT only'' concerning different training modes:

\begin{itemize}
  \item \textbf{FS}: training ``from scratch'' (with pre-trained ResNet18 weights only);
  \item \textbf{FT}: fine-tuning of a model trained with LS only.
\end{itemize}

Table~\ref{tab:train_mode} shows that, in the MVX$\rightarrow$WT setting, there was not much difference in using FS or FT as training modes for GVMD, with FT being slightly worse than FS regarding MODA and precision.
Both approaches outperformed the detector by Lima et al.~\cite{lima20223d} in this setting concerning MODA and precision.
In the WT$\rightarrow$MVX setting, ``GMVD FT'' was better than ``GMVD FS'' in all metrics.
In addition, ``GMVD FT'' outperformed the Lima et al.~\cite{lima20223d} detector in this setting regarding MODA, precision, and recall.

Next, we evaluated the use of multi-round automatic labeling.
In each round, we generate automatic labels and fine-tune the existing GMVD model using the newly labeled data.
We repeated this process for $3$ rounds, in which we used as training data ALT only in the first round and PLT only in the remaining rounds.
We perform pseudo-labeling using the GMVD model trained in the previous round.
Table~\ref{tab:multi_round} shows that the multi-round approach did not provide consistent improvements.
In the MVX$\rightarrow$WT setting, we got an increase in all metrics except recall after two rounds, but in the third round, all metrics except precision decreased.
In the WT$\rightarrow$MVX setting, MODA and recall got worse after each additional round.
We believe that the multi-round automatic labeling approach promotes overfitting.

\begin{figure*}[t!]
  \centering  
   \includegraphics[width=\linewidth]{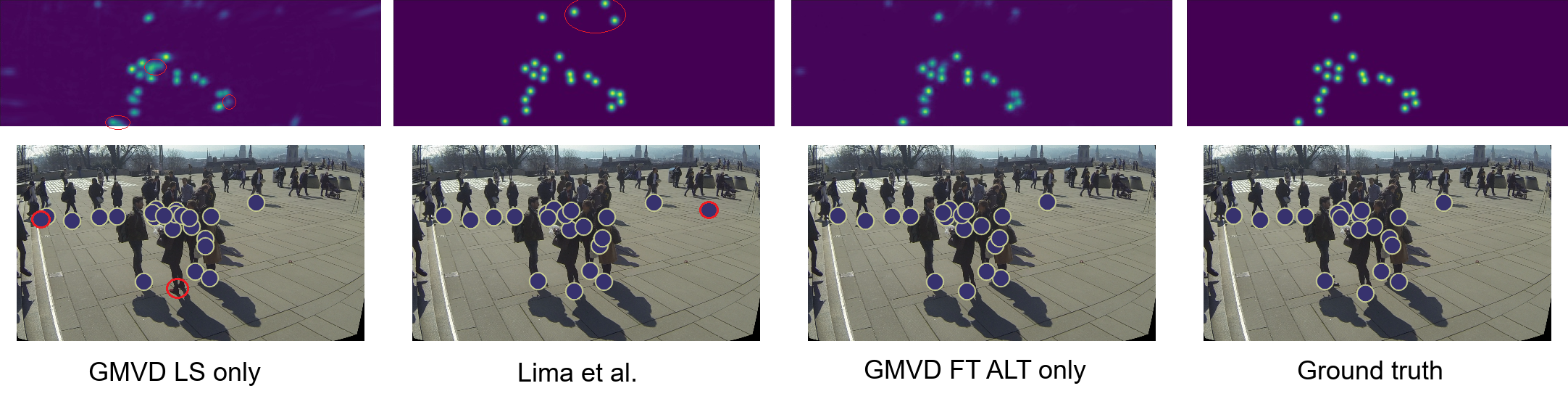}
   \caption{Results comparison for frame \#1885 of the WILDTRACK dataset.
   Top row: output occupancy heatmaps from each approach.
   Bottom row: 3D detections from each approach projected onto the image captured by camera \#3 (blue circles represent detected pedestrians).
   First column: results from GMVD~\cite{vora2023bringing} model trained with labeled MultiviewX source data only (``GMVD LS only'').
   Second column: results from the untrained detector by Lima et al.~\cite{lima20223d}.
   Third column: results from fine-tuning ``GMVD LS only'' using automatically labeled WILDTRACK data with the approach based on the detector by Lima et al.~\cite{lima20223d} (``GMVD FT ALT only'').
   Fourth column: ground truth.
   Major errors are highlighted in red.}
   \label{fig:qual_wt}
\end{figure*}

\begin{figure*}[t!]
  \centering  
   \includegraphics[width=\linewidth]{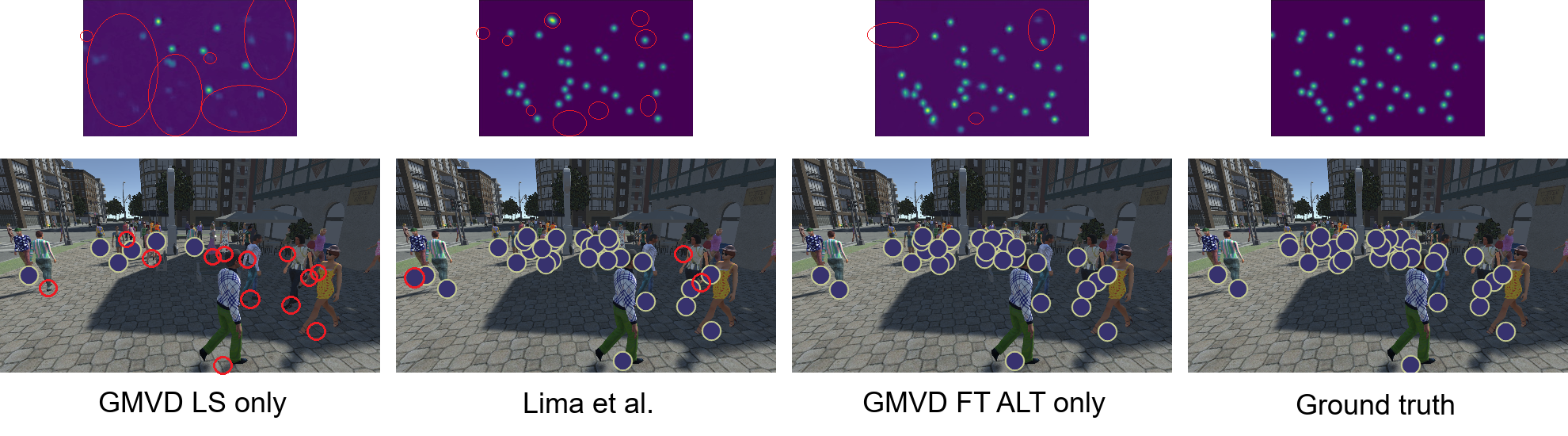}
   \caption{Results comparison for frame \#360 of the MultiviewX dataset.
   Top row: output occupancy heatmaps from each approach.
   Bottom row: 3D detections from each approach projected onto the image captured by camera \#6 (blue circles represent detected pedestrians).
   First column: results from GMVD~\cite{vora2023bringing} model trained with labeled WILDTRACK source data only (``GMVD LS only'').
   Second column: results from the untrained detector by Lima et al.~\cite{lima20223d}.
   Third column: results from fine-tuning ``GMVD LS only'' using automatically labeled MultiviewX data with the approach based on the detector by Lima et al.~\cite{lima20223d} (``GMVD FT ALT only'').
   Fourth column: ground truth.
   Major errors are highlighted in red.}
   \label{fig:qual_mvx}
\end{figure*}

\subsection{Qualitative evaluation}

We qualitatively compared three unlabeled approaches: ``GMVD FT ALT only'', ``GMVD LS only'', and the untrained detector by Lima et al.~\cite{lima20223d}.

Figure~\ref{fig:qual_wt} shows results obtained using them in a frame from WILDTRACK.
We can see that the output occupancy heatmap from ``GMVD LS only'' had some issues that resulted in the appearance of some false positives and one false negative.
We illustrate some of these problems in one camera image, where we can note one false positive relative to a pedestrian outside the area of interest and one false negative.
The detector by Lima et al.~\cite{lima20223d} presented three false positives, and we highlight one of them in the image from one of the cameras.
Such issues did not occur when we used GMVD with the FT ALT approach, leading to perfect detection results in the given frame.

Figure~\ref{fig:qual_mvx} depicts results obtained using the evaluated approaches in a frame from MultiviewX.
``GMVD LS only'' presented several false negatives, while the Lima et al.~\cite{lima20223d} detector obtained some false negatives and one false positive, as shown in the respective camera images.
In contrast, GMVD with the FT ALT approach presented no false positives and fewer false negatives.

\subsection{Limitations}

Like other multi-camera 3D pedestrian detection methods in the literature, the presented approaches are limited to estimating the 3D location of individuals on the ground plane.
Consequently, they cannot accurately determine the 3D position of people not standing on the ground, for example, when jumping.
As a result, their applicability may be restricted in domains such as dancing and sports analytics.

The PLT approach may not work well if there is a large generalization gap between source and target training data.
This issue is evidenced by the experiments for the WT$\rightarrow$MVX setting detailed in Table~\ref{tab:train_data}.

Regarding the ALT approach, it can inherit some limitations from the untrained detector used for generating new labels.
In our case, the detector by Lima et al.~\cite{lima20223d} may sometimes fail in scenes with severe occlusions and people wearing similar clothing.
Since we employ this detector for obtaining target labels, the model trained with such data may present similar problems.

\section{Conclusion}

We evaluated approaches for unlabeled multi-view 3D pedestrian detection that use generalizable AI methods based on automatic labeling.
The pseudo-labeling procedure improved the results of a supervised detector when it could generalize reasonably well to the unlabeled target dataset.
The automatic labeling approach based on an untrained detector and fine-tuning provided better results in all tested source$\rightarrow$target settings than directly using the untrained detector or a detector trained with labeled source data.
The multi-round automatic labeling procedure did not bring much improvement.

In future work, we intend to perform additional evaluations with more diverse datasets that promote less generalization gap, e.g., the GMVD dataset~\cite{vora2023bringing}.
We also plan to explore novel data augmentation approaches suitable to the multi-view 3D pedestrian detection problem~\cite{hou2021multiview, rui20223d, engilberge2023two}.
Regarding automatic labeling, we envisage using selective labeling~\cite{cho2022effective} and more sophisticated methods such as the Mean Teacher paradigm~\cite{kim2022unified}.
Finally, we intend to investigate the utilization of multi-view 3D tracking~\cite{nguyen2022lmgp} as an auxiliary task for self-supervised multi-view 3D pedestrian detection.


\section*{Acknowledgment}

The authors would like to thank CNPq (process 422728/2021-7) and JSPS (fellowship ID S22064) for partially funding this work.



\bibliographystyle{IEEEtran}
\bibliography{example}
%
%


\end{document}